# Gate Recurrent Unit Network based on Hilbert-Schmidt Independence Criterion for State-of-Health Estimation

Ziyue Huang, Lujuan Dang, Yuqing Xie, Wentao Ma, *Member, IEEE* and Badong Chen, *Senior Member, IEEE*

*Abstract*—State-of-health (SOH) estimation is a key step in ensuring the safe and reliable operation of batteries. Due to issues such as varying data distribution and sequence length in different cycles, most existing methods require health feature extraction technique, which can be time-consuming and labor-intensive. GRU can well solve this problem due to the simple structure and superior performance, receiving widespread attentions. However, redundant information still exists within the network and impacts the accuracy of SOH estimation. To address this issue, a new GRU network based on Hilbert-Schmidt Independence Criterion (GRU-HSIC) is proposed. First, a zero masking network is used to transform all battery data measured with varying lengths every cycle into sequences of the same length, while still retaining information about the original data size in each cycle. Second, the Hilbert-Schmidt Independence Criterion (HSIC) bottleneck, which evolved from Information Bottleneck (IB) theory, is extended to GRU to compress the information from hidden layers. To evaluate the proposed method, we conducted experiments on datasets from the Center for Advanced Life Cycle Engineering (CALCE) of the University of Maryland and NASA Ames Prognostics Center of Excellence. Experimental results demonstrate that our model achieves higher accuracy than other recurrent models.

## I. INTRODUCTION

THE increasing demand for energy puts significant pressure on both the environment and energy resources. As a result, there is a growing trend towards considering renewable energy as a substitute. Lithium-ion batteries have become widely used in electric vehicles (EVs), electric ships, and electric aircraft due to their high energy density, good power capability, and low self-discharge [1–3]. However, their electrochemical nature causes a gradual deterioration in battery health, leading to a degradation in power and capacity over time [4]. To ensure safe and reliable battery use, state-of-health (SOH) estimation techniques have been developed [5]. Currently, there are three main categories of state-of-health estimation methods: experiment or direct assessment approach, adaptive approach and data driven approach [6].

This work was supported by the National Key R&D Program of China (Grant No. 2021YFB2401904). (Corresponding author: Badong Chen.)

Ziyue Huang, Lujuan Dang, Yuqing Xie and Badong Chen are with the National Key Laboratory of Human-Machine Hybrid Augmented Intelligence, National Engineering Research Center for Visual Information and Applications, and Institute of Artificial Intelligence and Robotics, Xi'an Jiaotong University, Xi'an 710049, China (e-mail: zyhuang99@stu.xjtu.edu.cn; danglj@xjtu.edu.cn; felixxyq@stu.xjtu.edu.cn; chenbd@mail.xjtu.edu.cn;).

Wentao Ma is with the School of Electrical Engineering, Xi'an University of Technology, Xi'an 710048, China (e-mail: mawt@xaut.edu.cn)

Direct measurement methods obtain the capacity or resistance of the battery through a complete discharge after full charge. Common direct measurement methods include the coulomb counting method [7], the open circuit voltage method [8], and the electrochemical impedance spectroscopy method [9]. These methods are easy to collect data and calculate with high efficiency, but strict test conditions are required, making it difficult to apply them in practical scenarios. Model-based methods simulate the physical and chemical properties of batteries by applying battery data to construct electrochemical models such as the Thevenin battery model [10] and RC model [11, 12]. SOH is then estimated using partial differential equations (PDE) [13], Kalman filters [14, 15], and other filtering methods. These simple battery models combined with filtering methods can better simulate the dynamic characteristics of batteries and be adapted to more scenarios. However, the accuracy of these methods depends on the complexity and precision of the electrochemical model, and models constructed under different operating conditions are not applicable to each other.

With the advancement of artificial intelligence techniques, data-driven methods gain wide popularity in recent years, including machine learning methods such as SVM [16, 17], GPR [18], RF [19], and deep learning methods such as CNN [20], IRBFNN [21], LSTM [22], and GRU [23]. The data-driven models can extract information from the data and do not depend on the electrochemical model, which make them suitable for different types of batteries and operating conditions. However, most data-driven methods require health features extraction, as the length of battery data varies each cycle. This step captures health features, such as incremental calculation-based health features, time-based health features, envelope-area-based health features, and model-parameter-based health features [1], that reflect changes in the battery between cycles. Nonetheless, these methods tend to lose significant information, and the separation of health features extraction from subsequent model training increases the workload significantly.

A GRU network based on a sliding time window is proposed to estimate SOH [24], which does not require the pre-extraction of health features from battery data. However, this method only utilizes capacity data while ignoring other important data such as voltage, current, and temperature. In contrast, Fan et al. [25] effectively utilize voltage, current, and temperature data by applying the equally spaced resample method and feeding the data into a GRU-CNN network. Additionally, a kind of LSTM network with a masking network is proposed to handle variable-length time series data as input, eliminating the need for a specific input time size or length [26, 27]. As variants of RNN, LSTM and GRU excel in handling time-series tasks due to their recurrent connection, and address the issues of gradient explosion and gradient disappearance that arise in RNN. GRU and LSTM with masking network learn from all measured data and accurately estimate SOH with a simple process and concise structure, without requiring health feature extraction. Moreover, GRU has significantly fewer parameters than LSTM, yet demonstrates comparable performance and faster efficiency [28]. In certain scenarios, GRU even learns better from data than LSTM [29]. However, during the training of the aforementioned GRU network, there may exist redundancy in the learned information. To address this issue, the information bottleneck (IB) theory, which was originally proposed by Tishby [30], provides a promising approach to compress information. By extending the IB theory to the GRU, we can improve the performance of GRU with a simple structure.

The highlights of this paper are as follows:
- All battery data with varying lengths at each cycle are handled using a zero masking network. Our method integrates health feature extraction and model training.
- We propose a novel method that incorporates IB theory into GRU, which improves the ability of GRU to capture useful information.
- Our method is appropriate to constant and random charging/discharging data. The experiments on datasets from the CALCE of University of Marylan and NASA Random Walk are used to validate the superior performance of our method.

## II. METHOD

### A. Definition of State-of-Health

In Battery Management System (BMS), SOH is an indicator used to describe the state of health of the batteries, which is usually related to the power or maximum capacity of batteries. The definition of SOH is as in Eq. (1) here, and in our work, the target is to estimate the maximum capacity of each battery cycle.

$$SOH = \frac{C_{max}}{C_{nom}} \times 100\% \quad (1)$$

in which $C_{max}$ is the maximum capacity of the battery in the current cycle and $C_{new}$ is the nominal capacity of the fresh battery.

### B. Gate Recurrent Unit Network

As battery data has strong temporal properties, GRU can effectively capture information that helps estimate SOH. The inner structure of GRU is shown in Fig.1. Two gate loops - reset gate and updte gate are used to retain and desert information in GRU. The internal calculation of GRU is as follows:

$$r_t = \sigma(W_r x_t + U_r h_{t-1} + b_r) \quad (2)$$

$$z_t = \sigma(W_z x_t + U_z h_{t-1} + b_z) \quad (3)$$

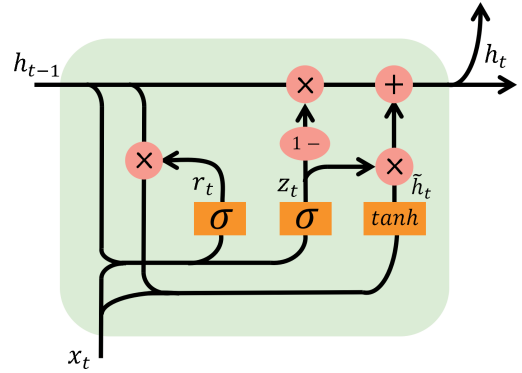

Fig. 1: Inner structure of GRU

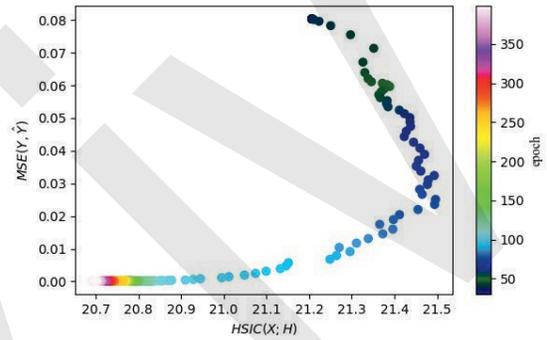

Fig. 2: Information plane of GRU with CS dataset

$$\tilde{h}_t = \tanh(W_h x_t + U_h(r_t \odot h_{t-1}) + b_h) \quad (4)$$

$$h_t = (1 - z_t) \odot h_{t-1} + z_t \odot \tilde{h}_t \quad (5)$$

where $x_t$ is the input of time $t$, $h_{t-1}$ is the previous state at time $t-1$, $r_t$ and $z_t$ are the outputs of reset gate and update gate, $\tilde{h}_t$ and $h_t$ are the candidate hidden state and hidden state at time $t$, $\odot$ represents the element-wise multiplication, $\sigma$ represents the sigmoid function, tanh represents the hyperbolic tangent function and $W_r$, $U_r$, $b_r$, $W_z$, $U_z$, $b_z$, $W_h$, $U_h$, $b_h$ are the parameters of the GRU model.

However, since lengths of values are different in each battery cycle, health feature extraction is required before putting the data into deep learning methods. But if zero masking network is added, GRU can directly learn from the battery data with different lengths in each cycle. The zero masking network converts the input sequence into a fixed-length sequence by adding zeros at the ends of sequences shorter than the longest one. The GRU layers then filter out all zeros at the end of sequences, while maintaining knowledge of the length of each cycle.

### C. Information Bottleneck and Hilbert-Schmidt Independence Criterion

The target of IB principle is to get a maximal compressed representation $T$ of the input $X$ while retaining $T$ with the maximum information about output Y. And it is shown in math as in Eq. (6).

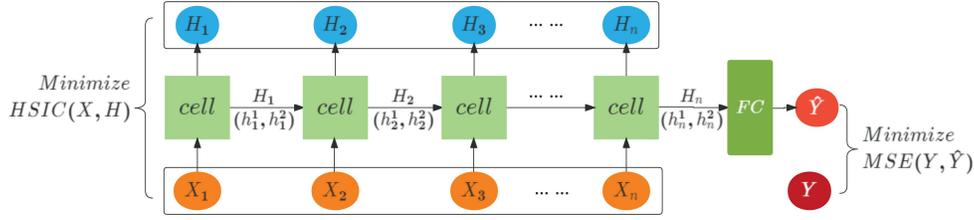

Fig. 3: GRU-HSIC structure

$$L_{IB} = I(T; Y) - \beta I(X; T) \quad (6)$$

where $I(\cdot; \cdot)$ denotes the mutual information and $\beta > 0$ is a Lagrange multiplier for measuring these two mutual information. For artificial neural network, $T$ could represent one of the intermediate layers. Because the input $X$, the information of hidden layers $H$ and the output $Y$ in the artificial neural network obviously form a Markov chain: $X \to H \to Y$. The operational mechanism of using IB to explain DNN [31] has been widely accepted. In particular, since the calculation process of IB is much more complex, we use the more computable Hilbert-Schmidt independence criterion (HSIC) bottleneck [32] which is evolved from IB instead of the mutual information between the hidden layer and the input. HSIC can measure the dependence of two random variables. Given two random variables $x$ and $y$, and two mapping functions $\phi : X \to F$ and $\varphi : Y \to G$. The formulation of HSIC [33–35] is as follows:

$$\begin{aligned} HSIC(X, Y) &= \|C_{XY}\|^2_{HS} \\ &= \|E[\phi(X)\varphi(Y)] - E[\phi(X)]E[\varphi(Y)]\|^2_{HS} \\ &= E[k(X, X')l(Y, Y')] + E[l(X, X')]E[l(Y, Y')] \\ &\quad - 2E[E[k(X, X')|X]E[l(Y, Y')|Y]] \end{aligned} \quad (7)$$

where $X'$ and $Y'$ are the independent identical distribution copy of $X$ and $Y$. $k(X, X')$ and $l(Y, Y')$ are Gaussian kernels in this article, and satisfy: $k(x, x') = \phi(x - x')$ and $l(y, y') = \varphi(y - y')$ for $x, x'$ in $R^p$ and $y, y'$ in $R^q$. $F$ and $G$ are the reproducing kernel Hilbert space (RKHS) and $C_{XY}$ is the cross-covariance operator. Given a sample $Z = (X_1, Y_1), ..., (X_n, Y_n)$ drawn from $P_{XY}$ which is the joint measure of $X$ and $Y$. As $k(X, X') = k(X', X)$ and $l(Y, Y') = l(Y', Y)$, we denote $k(X, X')$ as $k_X$, and denote $l(Y, Y')$ as $l_Y$. Therefore, the following form of HSIC can be obtained by the calculation of Eq. (8):

$$HSIC(X, Y) = \frac{1}{(n-1)^2} tr(k_X W k_Y W) \quad (8)$$

where $W = I_n - \frac{1}{n} \begin{pmatrix} 1 & \cdots & 1 \\ \vdots & \ddots & \vdots \\ 1 & \cdots & 1 \end{pmatrix}$.

To observe the changes in information while training GRU, we use the HSIC between the hidden layer and input, and Mean Squared Error (MSE) between the model output and truth, instead of mutual information. As shown in Figure 2, the HSIC between the input and hidden layers initially increases and then decreases, indicating that there is redundancy while training. Considering that applying the IB theory can effectively reduce redundancy and enhance the performance and efficiency of GRU, we construct a new GRU model with HSIC bottleneck named GRU-HSIC. The structure is shown in Fig.3 and the loss function is given in Eq. (9).

$$L_{GRU-HSIC} = MSE(Y, \hat{Y}) + \beta HSIC(X; H) \quad (9)$$

where $Y$ presents the truth of the maximize capacity, $\hat{Y}$ represents the estimation of model, $X$ means the input, $H$ stands for the information of hidden layers in model and $\beta$ is a Lagrange multiplier.

III. EXPERIMENT

A. Datasets

Three datasets are selected in our experiment. Two are from the Center for Advanced Life Cycle Engineering (CALCE) of the University of Maryland [36], one is from NASA Ames Prognostics Center of Excellence [37, 38].

Eight batteries are selected from CS and CX2 datasets of CALCE respectively, namely CS2_35, CS2_36, CS2_37, CS2_38 and CX2_34, CX2_36, CX2_37, CX2_38. These batteries have the same polar materials - $LiCoO_2$, the same constant current-constant voltage (CC-CV) charging process and constant current (CC) discharging process, but the capacity rating of the CS battery is 1100 mAh while 1350 mAh for CX2 battery.

Four random walk discharging batteries (RW13, RW14, RW15, RW16) from NASA Ames Prognostics Center of Excellence are chosen here. They are continuously operated by repeatedly charging them to 4.2V and then discharging them to 3.2V with random current between 0.5A and 5A. Reference charging and discharging cycles are carried out after every 50 cycles to provide benchmarks for battery state health.

The evaluation criteria used here is *MSE*, *RMSE*, *MAE*, *MAPE*, *SMAPE* and *relative erros* which are defined as Eqs. (10 - 15).

$$MSE = \frac{1}{n} \sum_{i=1}^{n} (y_i - \hat{y}_i)^2 \quad (10)$$

$$RMSE = \sqrt{\frac{1}{n} \sum_{i=1}^{n} (y_i - \hat{y}_i)^2} \quad (11)$$

$$MAE = \frac{\sum_{i=1}^{n} |y_i - \hat{y}_i|}{n} \quad (12)$$

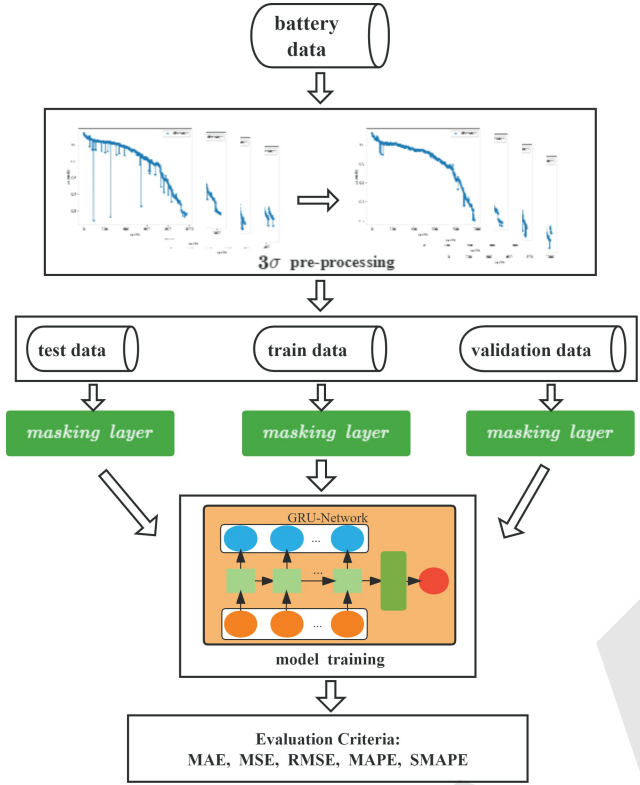

Fig. 4: Flow chart

TABLE I: setting of models in CS dataset

| model | hidden states | net layers | seed | learning rate | β | epoch |
|---|---|---|---|---|---|---|
| RNN | 2 | 3 | 13 | 0.01 | / | 900 |
| LSTM | 2 | 3 | 13 | 0.01 | / | 900 |
| GRU | 2 | 3 | 13 | 0.005 | / | 900 |
| GRU-HSIC | 2 | 3 | 13 | 0.005 | 0.001 | 900 |

$$MAPE = \frac{100\%}{n} \sum_{i=1}^{n} \left| \frac{y_i - \hat{y}_i}{y_i} \right| \tag{13}$$

$$SMAPE = \frac{100\%}{n} \sum_{i=1}^{n} \frac{2|y_i - \hat{y}_i|}{(|y_i| + |\hat{y}_i|)} \tag{14}$$

$$relative\ errors = |y_i - \hat{y}_i| \tag{15}$$

Experiments are carried on four NVIDIA 3090 GPUs with Python 3.7.0.

TABLE II: MAE score compare in CS dataset

| Model | Average MAE | MAE in CS2_35 | MAE in CS2_36 | MAE in CS2_37 | MAE in CS2_38 |
|---|---|---|---|---|---|
| THF-RF | 0.015229 | 0.008943 | 0.031082 | 0.009837 | 0.011056 |
| THF-GPR | 0.026747 | 0.023732 | 0.038745 | 0.021040 | 0.023472 |
| RNN | 0.071835 | 0.050642 | 0.081031 | 0.081577 | 0.074091 |
| LSTM | 0.024486 | 0.017969 | 0.025427 | 0.016974 | 0.037571 |
| GRU | 0.009695 | 0.010174 | 0.013220 | 0.009949 | 0.005437 |
| GRU-HSIC | **0.004321** | **0.004553** | **0.006372** | **0.003656** | **0.002704** |

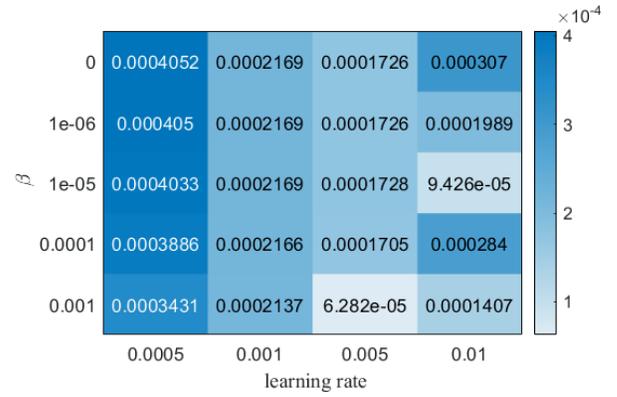

Fig. 5: MSE score comparion of GRU and GRU-HSIC in CS dataset

TABLE III: RMSE score compare in CS dataset

| Model | Average RMSE | RMSE in CS2_35 | RMSE in CS2_36 | RMSE in CS2_37 | RMSE in CS2_38 |
|---|---|---|---|---|---|
| THF-RF | 0.053642 | 0.035144 | 0.119698 | 0.036704 | 0.023021 |
| THF-GPR | 0.053850 | 0.041323 | 0.064588 | 0.052288 | 0.057202 |
| RNN | 0.101645 | 0.068206 | 0.133452 | 0.102226 | 0.102698 |
| LSTM | 0.041674 | 0.023447 | 0.042906 | 0.020965 | 0.079379 |
| GRU | 0.012250 | 0.010843 | 0.020242 | 0.010098 | 0.007815 |
| GRU-HSIC | **0.006982** | **0.005389** | **0.013404** | **0.005224** | **0.003911** |

### B. Experiment in CS Dataset

For the CS dataset, the model's ground truth is obtained by integrating the discharging and charging current data. Since there are more outliers in the charging part, priority is given to the discharging parts during the experiments to obtain an accurate approximation of the ground truth. The entire time series' voltage, current, and resistance (V, I, R) are used as the model input, and the corresponding capacity after outlier handling in each cycle is used as the label.

In training process, we use two batteries as the training data, one battery as the verify data and the rest one battery as the test data. For example, the batteries CS27 and CS2_38 are used as the training data, the battery CS2_36 is used as the verify data and the rest battery CS2_35 is used as the test data in the experiment of battery CS2_35. For the RNN, LSTM, GRU, and GRU-HSIC, all voltage, current, and resistance data are used, resulting in high complexity of the recurrent network. To avoid difficulties in applying the model due to its large size, we focus on models with simple architectures. The model configurations for these four models are shown in Table I and we choose the best results of all four models. As an example of GRU and GRU-HSIC, we plot the heatmap of experiments with different parameters (learning rate and β) in Fig.5 where the best parameters are chosen when learning rate is 0.005 and beta is 0.0 and 0.001. In addition, we also used the random forest (RF) model and the Gaussian process regression (GPR) model with classical time-based health features (THF) as a comparison.

From the Table II and III, we can find that the best MAE and RMSE of batteries appear in GRU-HSIC. And from Fig.7, it is obviously that the accuracy of GRU-HSIC is much higher than RNN and LSTM and slightly higher than GRU in every cycle of battery life. And in the last few hundred cycles of all four batteries and the first 200 cycles of battery CS2_38, there are significant fluctuations in RNN and LSTM, but GRU and GRU-HSIC still retain good performance. The errors of GRU-HSIC basically fluctuate within 0.015. The accuracy of THF-RF and THF-GPR is similar to LSTM, but the overall precision of THF-RF is lower, and the precision of THF-GPR is unstable across these four batteries. Compared to these two methods with time-based health features, GRU and GRU-HSIC perform much better. This means that both GRU and GRU-HSIC have good robustness and can obtain the global optimal solution very well. And the worst performance of RNN may be attributed to its overly simple recurrent cell structure. The poor performance of LSTM may be contributed to its retaining too much redundant information during training. But GRU is able to sieve out unwanted information due to its simplified structure, and the addition of the HSIC bottleneck structure further filters the information of hidden layers. From Fig.6, GRU-HSIC maintains optimal performance under all five evaluation criteria (MAE, MSE, RMSE, MAPE, SMAPE) and has significant advantages.

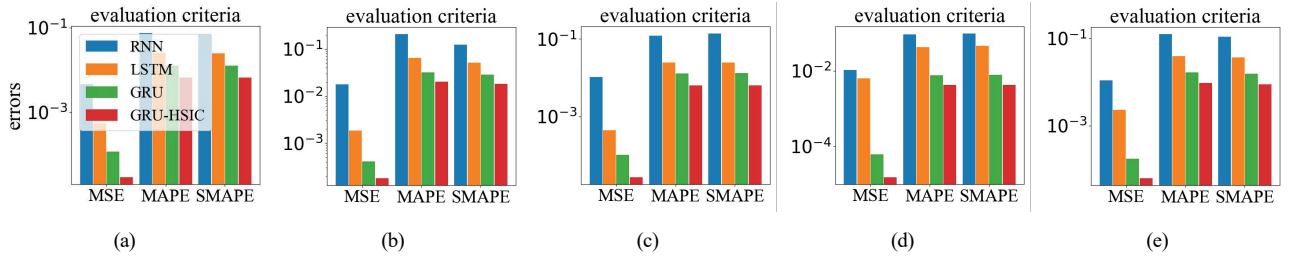

Fig. 6: Evaluation in CS dataset: (**a**) CS2_35, (**b**) CS2_36, (**c**) CS2_37,(**d**) CS2_38, (**e**) the average evaluatin.

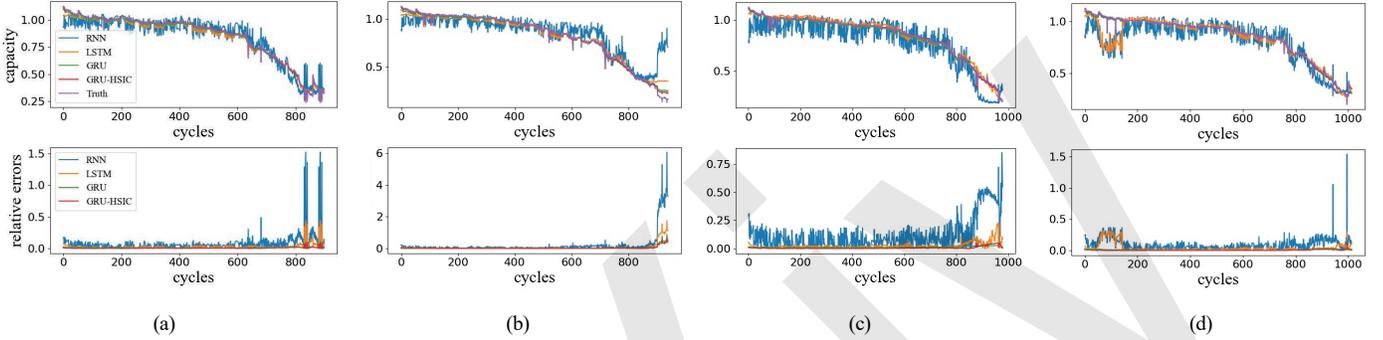

Fig. 7: Capacity and relative errors in CS dataset: (**a**) CS2_35, (**b**) CS2_36, (**c**) CS2_37,(**d**) CS2_38.

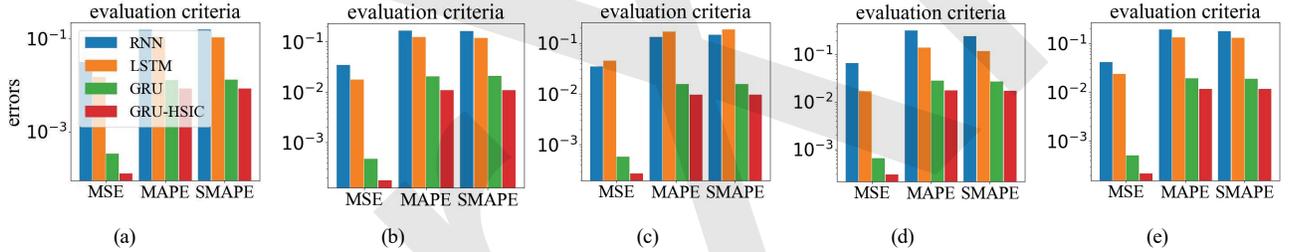

Fig. 8: Evaluation in CX2 datset: (**a**) CX2_34, (**b**) CX2_36, (**c**) CX2_37,(**d**) CX2_38, (**e**) the average evaluatin.

TABLE IV: setting of models in CX2 dataset

| model | hidden states | net layers | seed | learning rate | $\beta$ | epoch |
|---|---|---|---|---|---|---|
| RNN | 2 | 1 | 11 | 0.01 | / | 900 |
| LSTM | 2 | 1 | 11 | 0.005 | / | 900 |
| GRU | 2 | 1 | 11 | 0.005 | / | 900 |
| GRU-HSIC | 2 | 1 | 11 | 0.005 | 0.001 | 900 |

TABLE V: MAE score compare in CX2 dataset

| Model | Average MAE | MAE in CX2_34 | MAE in CX2_36 | MAE in CX2_37 | MAE in CX2_38 |
|---|---|---|---|---|---|
| THF-RF | 0.028219 | 0.022766 | 0.012403 | **0.007014** | 0.070692 |
| THF-GPR | 0.023852 | 0.016552 | 0.015625 | 0.018064 | 0.045166 |
| RNN | 0.167329 | 0.153342 | 0.157970 | 0.155531 | 0.202474 |
| LSTM | 0.123287 | 0.102471 | 0.112975 | 0.189748 | 0.087954 |
| GRU | 0.017001 | 0.012931 | 0.018360 | 0.017600 | 0.019114 |
| GRU-HSIC | **0.011033** | **0.008737** | **0.010892** | 0.011177 | **0.013325** |

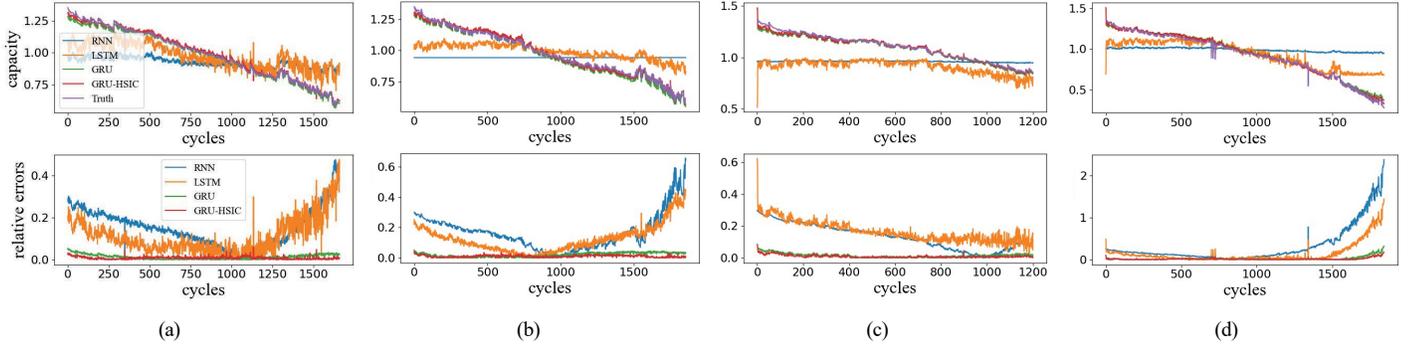

Fig. 9: Capacity and relative errors in CX2 dataset: (**a**) CX2_34, (**b**) CX2_36, (**c**) CX2_37, (**d**) CX2_38.

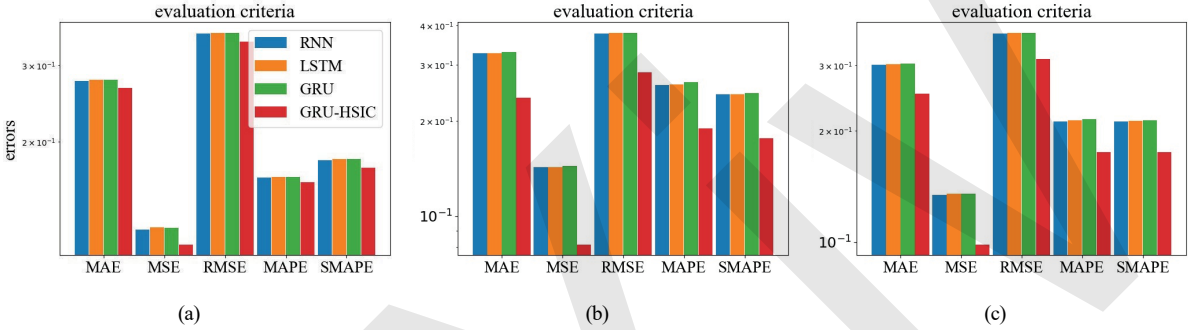

Fig. 10: Evaluation in NASA RW dataset: (**a**) the first experiment, (**b**) the second experiment, (**c**) the average evaluatin.

TABLE VI: RMSE score compare in CX2 dataset

| Model | Average RMSE | RMSE in CX2_34 | RMSE in CX2_36 | RMSE in CX2_37 | RMSE in CX2_38 |
| --- | --- | --- | --- | --- | --- |
| THF-RF | 0.066520 | 0.041006 | 0.033559 | 0.017917 | 0.173600 |
| THF-GPR | 0.048207 | 0.034264 | 0.037784 | 0.038850 | 0.081931 |
| RNN | 0.201727 | 0.178027 | 0.186825 | 0.186317 | 0.255737 |
| LSTM | 0.149096 | 0.122604 | 0.133215 | 0.211353 | 0.129210 |
| GRU | 0.022353 | 0.018316 | 0.021702 | 0.024103 | 0.025292 |
| GRU-HSIC | **0.014502** | **0.011191** | **0.013266** | **0.016436** | **0.017113** |

TABLE VII: setting of models in NASA RW dataset

| model | hidden states | net layers | seed | learning rate | $\beta$ | epoch |
| --- | --- | --- | --- | --- | --- | --- |
| RNN | 2 | 3 | 12 | 0.05 | / | 1000 |
| LSTM | 2 | 3 | 12 | 0.05 | / | 1000 |
| GRU | 2 | 3 | 12 | 0.05 | / | 1000 |
| GRU-HSIC | 2 | 3 | 12 | 0.05 | 2.0 | 1000 |

*C. Experiment in CX2 Dataset*

The charging parts in CX2 datasets also contain more outliers compared to the discharging parts, so the discharging parts are preferred in experiments. The training and test set in CX2 datasets is the same as those in CS datasets. The model set is also similar to that in CS datasets and is shown in Table IV, and we chose the best parameters for those structures. The final results are presented in Fig.9, Fig.8, Table V and Table VI. And the MAE and the RMSE of battery CX2_34 - CX2_38 in GRU-HSIC are the best. The performance of THF-RF and THF-GPR is better than LSTM but worse than GRU and GRU-HSIC, among which THF-RF performs very well on battery CX2_37 but poorly on battery CX2_38. From Fig.9, although the model structure becomes simpler, GRU-HSIC still maintains good accuracy among all these recurrent models and the relative errors also basically fluctuate within 0.02. However, the performance of LSTM and RNN model becomes pretty bad. The trend that LSTM follows the truth is more obvious that RNN, but the fluctuation of LSTM is too large. GRU and GRU-HSIC perform very well during all the cycles compared to LSTM and RNN. And from Fig.8, GRU-HSIC also maintains optimal performance under all five evaluation criteria (MAE, MSE, RMSE, MAPE, SMAPE) in CX2 datasets. It is verified that simpler structure of GRU-HSIC also has good performance.

*D. Experiment in NASA-RW Dataset*

In NASA RW dataset, two standard charging and discharging processes serve as references between every two random charging and discharging experiments. The truth of the maximum capacity for estimating SOH is obtained by integrating the current from the second discharging process in each interval. The random parts' voltage, current, and temperature (V, I, T) are used as inputs for the model. Two experiments are conducted using the NASA RW dataset. The first experiment uses RW15 and RW16 as training data, RW13 as testing data, and RW14 as validation data. The second experiment uses RW13 and RW16 as training data, RW14 as testing data, and RW15 as validation data. Parameter setting is 4shown in Table VII. In addition, we compare these four models where the learning rates are the same this time. Due to the difficulty of extracting health features from random data, we do not use health-feature-based methods as a comparison here. The final results are shown in Fig.10, where the performance of GRU-HSIC is far superior to the other three models in all five evaluation criteria (MAE, MSE, RMSE, MAPE, SMAPE). We also observe that RNN, LSTM and GRU perform similarly poorly, possibly because training a network with such completely random distributed datasets is challenging. However, GRU-HSIC performs well, possibly due to HSIC bottleneck's compression ability which allowed it to obtain critical and useful information from the random data.

## IV. CONCLUSION

In this paper we develop a new GRU network based on Hilbert-Schmidt Independence Criterion (GRU-HSIC) for SOH estiamtion. This method simplifies the steps of SOH estimation as it combines health feature extraction and model training together. Information is acquired without being influenced by human perception, thereby reducing the possibility of information omission and minimizing manual subjective operations. The HSIC bottleneck developed from IB theory is extend to GRU for estimating SOH. All the battery data can be directly put into the model through a zero masking network, and the HSIC bottleneck help optimizing the process of model training well. The validation results on different constant and random charging and discharging datasets effectively indicate that the GRU-HSIC model outperforms the time-based data-driven model, as well as the RNN, LSTM, and GRU models which does not require health feature extraction. And the experiments on NASA RW dataset demonstrate that GRU-HSIC is not limited to regular data from the laboratory, but it is also effective for data with a heterogeneous distribution.

## REFERENCES

[1] X. Shu, S. Shen, J. Shen, Y. Zhang, G. Li, Z. Chen, and Y. Liu, "State of health prediction of lithium-ion batteries based on machine learning: Advances and perspectives," *Iscience*, vol. 24, no. 11, p. 103265, 2021.

[2] E. Vanem, C. B. Salucci, A. Bakdi, and Ø. Å. sheim Alnes, "Data-driven state of health modelling—a review of state of the art and reflections on applications for maritime battery systems," *Journal of Energy Storage*, vol. 43, p. 103158, 2021.

[3] Z. Deng, X. Lin, J. Cai, and X. Hu, "Battery health estimation with degradation pattern recognition and transfer learning," *Journal of Power Sources*, vol. 525, p. 231027, 2022.

[4] J. Tian, R. Xiong, W. Shen, and F. Sun, "Electrode ageing estimation and open circuit voltage reconstruction for lithium ion batteries," *Energy Storage Materials*, vol. 37, pp. 283–295, 2021.

[5] Y. Li, K. Liu, A. M. Foley, A. Zülke, M. Berecibar, E. Nanini-Maury, J. Van Mierlo, and H. E. Hoster, "Data-driven health estimation and lifetime prediction of lithium-ion batteries: A review," *Renewable and sustainable energy reviews*, vol. 113, p. 109254, 2019.

[6] M. H. Lipu, M. Hannan, A. Hussain, M. Hoque, P. J. Ker, M. M. Saad, and A. Ayob, "A review of state of health and remaining useful life estimation methods for lithium-ion battery in electric vehicles: Challenges and recommendations," *Journal of cleaner production*, vol. 205, pp. 115–133, 2018.

[7] K. S. Ng, C.-S. Moo, Y.-P. Chen, and Y.-C. Hsieh, "Enhanced coulomb counting method for estimating state-of-charge and state-of-health of lithium-ion batteries," *Applied energy*, vol. 86, no. 9, pp. 1506–1511, 2009.

[8] Y. Cui, P. Zuo, C. Du, Y. Gao, J. Yang, X. Cheng, Y. Ma, and G. Yin, "State of health diagnosis model for lithium ion batteries based on real-time impedance and open circuit voltage parameters identification method," *Energy*, vol. 144, pp. 647–656, 2018.

[9] Y. Fu, J. Xu, M. Shi, and X. Mei, "A fast impedance calculation-based battery state-of-health estimation method," *IEEE Transactions on Industrial Electronics*, vol. 69, no. 7, pp. 7019–7028, 2021.

[10] P. A. Topan, M. N. Ramadan, G. Fathoni, A. I. Cahyadi, and O. Wahyunggoro, "State of charge (soc) and state of health (soh) estimation on lithium polymer battery via kalman filter," in *2016 2nd International Conference on Science and Technology-Computer (ICST)*. IEEE, 2016, pp. 93–96.

[11] L. Fang, J. Li, and B. Peng, "Online estimation and error analysis of both soc and soh of lithium-ion battery based on dekf method," *Energy Procedia*, vol. 158, pp. 3008–3013, 2019.

[12] M. Gholizadeh and A. Yazdizadeh, "Systematic mixed adaptive observer and ekf approach to estimate soc and soh of lithium–ion battery," *IET Electrical Systems in Transportation*, vol. 10, no. 2, pp. 135–143, 2020.

[13] S. J. Moura, N. A. Chaturvedi, and M. Krstic´, "Pde estimation techniques for advanced battery management systems—part ii: Soh identification," in *2012 American Control Conference (ACC)*. IEEE, 2012, pp. 566–571.

[14] X. Li, Z. Huang, J. Tian, and Y. Tian, "State-of-charge estimation tolerant of battery aging based on a physics-based model and an adaptive cubature kalman filter," *Energy*, vol. 220, p. 119767, 2021.

[15] W. Yan, B. Zhang, G. Zhao, S. Tang, G. Niu, and X. Wang, "A battery management system with a lebesgue-sampling-based extended kalman filter," *IEEE transactions on industrial electronics*, vol. 66, no. 4, pp. 3227–3236, 2018.

[16] X. Feng, C. Weng, X. He, X. Han, L. Lu, D. Ren, and M. Ouyang, "Online state-of-health estimation for li-ion battery using partial charging segment based on support vector machine," *IEEE Transactions on Vehicular Technology*, vol. 68, no. 9, pp. 8583–8592, 2019.

[17] J. Meng, L. Cai, G. Luo, D.-I. Stroe, and R. Teodorescu, "Lithium-ion battery state of health estimation with short-term current pulse test and support vector machine," *Microelectronics Reliability*, vol. 88, pp. 1216–1220, 2018.
7

[18] H. Huang, J. Meng, Y. Wang, F. Feng, L. Cai, J. Peng, and T. Liu, "A comprehensively optimized lithium-ion battery state-of-health estimator based on local coulomb counting curve," *Applied Energy*, vol. 322, p. 119469, 2022.

[19] K. Liu, X. Hu, H. Zhou, L. Tong, W. D. Widanage, and J. Marco, "Feature analyses and modeling of lithium-ion battery manufacturing based on random forest classification," *IEEE/ASME Transactions on Mechatronics*, vol. 26, no. 6, pp. 2944–2955, 2021.

[20] G. Lee, D. Kwon, and C. Lee, "A convolutional neural network model for soh estimation of li-ion batteries with physical interpretability," *Mechanical Systems and Signal Processing*, vol. 188, p. 110004, 2023.

[21] J. Wu, L. Fang, G. Dong, and M. Lin, "State of health estimation of lithium-ion battery with improved radial basis function neural network," *Energy*, vol. 262, p. 125380, 2023.

[22] H. Sun, J. Sun, K. Zhao, L. Wang, and K. Wang, "Data-driven ica-bi-lstm-combined lithium battery soh estimation," *Mathematical Problems in Engineering*, vol. 2022, pp. 1–8, 2022.

[23] H. Sun, D. Yang, J. Du, P. Li, and K. Wang, "Prediction of li-ion battery state of health based on data-driven algorithm," *Energy Reports*, vol. 8, pp. 442–449, 2022.

[24] L. Ungurean, M. V. Micea, and G. Cârstoiu, "Online state of health prediction method for lithium-ion batteries, based on gated recurrent unit neural networks," *International journal of energy research*, vol. 44, no. 8, pp. 6767–6777, 2020.

[25] Y. Fan, F. Xiao, C. Li, G. Yang, and X. Tang, "A novel deep learning framework for state of health estimation of lithium-ion battery," *Journal of Energy Storage*, vol. 32, p. 101741, 2020.

[26] J. Hong, Z. Wang, W. Chen, L. Wang, P. Lin, and C. Qu, "Online accurate state of health estimation for battery systems on real-world electric vehicles with variable driving conditions considered," *Journal of Cleaner Production*, vol. 294, p. 125814, 2021.

[27] W. Li, N. Sengupta, P. Dechent, D. Howey, A. Annaswamy, and D. U. Sauer, "Online capacity estimation of lithium-ion batteries with deep long short-term memory networks," *Journal of power sources*, vol. 482, p. 228863, 2021.

[28] Y.-X. Wang, Z. Chen, and W. Zhang, "Lithium-ion battery state-of-charge estimation for small target sample sets using the improved gru-based transfer learning," *Energy*, vol. 244, p. 123178, 2022.

[29] J. Chung, C. Gulcehre, K. Cho, and Y. Bengio, "Empirical evaluation of gated recurrent neural networks on sequence modeling," *arXiv preprint arXiv:1412.3555*, 2014.

[30] N. Tishby, F. C. Pereira, and W. Bialek, "The information bottleneck method," *arXiv preprint physics/0004057*, 2000.

[31] N. Tishby and N. Zaslavsky, "Deep learning and the information bottleneck principle," in *2015 ieee information theory workshop (itw)*. IEEE, 2015, pp. 1–5.

[32] W.-D. K. Ma, J. Lewis, and W. B. Kleijn, "The hsic bottleneck: Deep learning without back-propagation," in *Proceedings of the AAAI conference on artificial intelligence*, vol. 34, no. 04, 2020, pp. 5085–5092.

[33] A. Gretton, O. Bousquet, A. Smola, and B. Schölkopf, "Measuring statistical dependence with hilbert-schmidt norms," in *Algorithmic Learning Theory: 16th International Conference, ALT 2005, Singapore, October 8-11, 2005. Proceedings 16*. Springer, 2005, pp. 63–77.

[34] A. Meynaoui, M. Albert, B. Laurent, and A. Marrel, "Adaptive test of independence based on hsic measures," 2019.

[35] Y. He, D. Yan, W. Xie, Y. Zhang, Q. He, and Y. Yang, "Optimizing graph neural network with multiaspect hilbert-schmidt independence criterion," *IEEE Transactions on Neural Networks and Learning Systems*, 2022.

[36] B. Bole, C. Kulkarni, and M. Daigle, "Randomized battery usage data set, nasa ames prognostics data repository. nasa ames research center, moffett field, ca," 2014.

[37] W. He, N. Williard, M. Osterman, and M. Pecht, "Prognostics of lithium-ion batteries based on dempster–shafer theory and the bayesian monte carlo method," *Journal of Power Sources*, vol. 196, no. 23, pp. 10 314–10 321, 2011.

[38] Y. Xing, E. W. Ma, K.-L. Tsui, and M. Pecht, "An ensemble model for predicting the remaining useful performance of lithium-ion batteries," *Microelectronics Reliability*, vol. 53, no. 6, pp. 811–820, 2013.